\documentclass[final,12pt]{clear2023} 

\newtheorem{defin}{Definition}
\newtheorem{prop}{Proposition}
\newtheorem{corollary}{Corollary}
\DeclareMathOperator{\doop}{\textit{do}}
\newcommand{\scope}[1]{\mathbf{sc}(#1)}
\DeclareMathOperator{\pa}{pa}
\DeclareMathOperator{\Pa}{Pa}
\DeclareMathOperator{\ch}{ch}

\newcommand{\inner}[1]{\langle #1 \rangle}
\usepackage[shortlabels]{enumitem}

\usepackage{relsize}
\usepackage{tikz} 
\usepackage{blkarray}
\usepackage{algorithmic}
\usepackage{adjustbox}
\usepackage{tabularx}
\usepackage{multirow}
\usepackage{pifont}
\newcommand*\colourcheck[1]{%
  \expandafter\newcommand\csname #1check\endcsname{\textcolor{#1}{\ding{52}}}%
}
\colourcheck{blue}
\colourcheck{green}
\colourcheck{red}
\newcommand*\colourcross[1]{%
  \expandafter\newcommand\csname #1cross\endcsname{\textcolor{#1}{\ding{55}}}%
}
\colourcross{blue}
\colourcross{green}
\colourcross{red}
\usepackage{array}
\usepackage{siunitx,booktabs}
\usepackage{capt-of}
\usepackage[most]{tcolorbox}
\tcbset{textmarker/.style={%
        enhanced,
        parbox=false,boxrule=0mm,boxsep=0mm,arc=0mm,
        outer arc=0mm,left=6mm,right=3mm,top=7pt,bottom=7pt,
        toptitle=1mm,bottomtitle=1mm,oversize}}
\newtcolorbox{hintBox}{textmarker,
    borderline west={6pt}{0pt}{yellow},
    colback=yellow!10!white}

\title[Inference in Causal Models]{A Taxonomy for Inference in Causal Model Families}
\usepackage{times}

\author{
\textbf{Matej Zečević}\textsuperscript{\rm 1}
\quad \textbf{Devendra Singh Dhami}\textsuperscript{\rm 1,3} \quad \textbf{Kristian Kersting}\textsuperscript{\rm 1-4}\\
{\normalfont \textsuperscript{\rm 1}Computer Science Department, TU Darmstadt, \textsuperscript{\rm 2}Centre for Cognitive Science, TU Darmstadt, \textsuperscript{\rm 3}Hessian Center for AI (hessian.AI),
\textsuperscript{\rm 4}DFKI}
}

\begin{document}

\maketitle

\begin{abstract}
  Neurally-parameterized Structural Causal Models in the Pearlian notion to causality, referred to as NCM, were recently introduced as a step towards next-generation learning systems. However, said NCM are only concerned with the learning aspect of causal inference but totally miss out on the architecture aspect. That is, actual causal inference within NCM is intractable in that the NCM won't return an answer to a query in polynomial time. This insight follows as corollary to the more general statement on the intractability of arbitrary SCM parameterizations, which we prove in this work through classical 3-SAT reduction. Since future learning algorithms will be required to deal with both high dimensional data and highly complex mechanisms governing the data, we ultimately believe work on tractable inference for causality to be decisive. We also show that not all ``causal'' models are created equal. More specifically, there are models capable of answering causal queries that are not SCM, which we refer to as \emph{partially causal models} (PCM). We provide a tabular taxonomy in terms of tractability properties for all of the different model families, namely correlation-based, PCM and SCM. To conclude our work, we also provide some initial ideas on how to overcome parts of the intractability of causal inference with SCM by showing an example of how parameterizing an SCM with SPN modules can at least allow for tractable mechanisms.
  
  We hope that our impossibility result alongside the taxonomy for tractability in causal models can raise awareness for this novel research direction since achieving success with causality in real world downstream tasks will not only depend on learning correct models as we also require having the practical ability to gain access to model inferences.
\end{abstract}


\section{Introduction}
Causal interactions stand at the center of human cognition thus being of high value to science, engineering, business, and law \citep{penn2007causal}. Questions like ``What if?'' and ``Why?'' were discovered to be central to how children perform exploration, as recent strides in developmental psychology suggest \citep{gopnik2012scientific, buchsbaum2012power, pearl2018book}, and similar to the scientific method. Whereas artificial intelligence research dreams of an automatation to the scientist's manner \citep{mccarthy1998artificial, mccarthy1981some, steinruecken2019automatic}. Deep learning's advance brought universality in approximation i.e., for any function there will exist a neural network that is close in approximation to arbitrary precision \citep{cybenko1989approximation, hornik1991approximation}. The field has seen tremendous progress ever since, see for instance \citep{krizhevsky2012imagenet, mnih2013playing,vaswani2017attention}. Thereby, the integration of causality with deep learhing is crucial for achieving human-level intelligence. Preliminary attempts, for so-called neural-causal models \citep{xia2021causal, pawlowski2020deep, zevcevic2021interventional} suggest to be promising.

While causality has been thoroughly formalized within the last decade \citep{pearl2009causality,peters2017elements}, and deep learning advanced, the issue of tractability of inference \citep{cooper1990computational, roth1996hardness, choi2020probabilistic} has been left unscathed. It is generally known that semantic graphs like Bayesian Networks (BNs, \citet{pearl1995bayesian}) scale exponentially for marginal inference, while computation graphs (or probabilistic circuits) like sum-product networks (SPNs, \citet{poon2011sum}) scale in polynomial (if not linear) time. A conversion method developed by \citet{zhao2015relationship} showed how to compile back and forth between SPNs and BNs. Yet, diverging views on tractable causal inference were reported, see discussions by \cite{papantonis2020interventions} or \citet{zevcevic2021interventional}. The former argues using the aforementioned conversion scheme, which leads to a degenerate BN with no causal semantics, while the latter proposes a partial neural-causal model that leverages existing interventional data to perform tractable causal inferences. Motivated by these discrepancies and lack of clarity, this work focusses on investigating systematically if, when, how and also under what cost the different types of causal inference occur in tractable manner. Our investigation of causal models reveals a tabular taxonomy to summarize recent research efforts and clarify what should be subject of further research, but also reveals a newly proposed model as an initial step towards said goal.

We make the following contributions: (1) We prove the general impossibility result of tractable inference within parameterized SCM, (2) we provide a comprehensive view onto the different trade-offs between expressivity of a causal model and tractability in inference, classifying our models along with their properties within a tabular taxonomy, and finally (3) based on our taxonomy we propose a new model called TNCM that can perform linear time mechanism inference opposed to higher degree polynomials as in NCM \citep{xia2021causal}.

We make our code repository with TNCM and visualizations publicly available at: \url{https://anonymous.4open.science/r/TNCM/}

\section{Brief Overview on Background and Related Work} \label{sec:two}
Let us briefly review the background on both key concepts from causality and the main tractable model class of concern, sum-product networks (SPNs). Because SPNs will play a central role in the discussion of this paper, since they take a singular role in model families that are truly tractable, we refer readers unfamiliar with the model family to the overview provided by \citet{paris2020sum}.

{\bf Causal Inference.}
Following the Pearlian notion of Causality \citep{pearl2009causality}, an SCM is defined as a 4-tuple $\mathcal{M}:=\inner{\mathbf{U},\mathbf{V},\mathcal{F},P(\mathbf{U})}$ where the so-called structural equations
\begin{align}\label{eq:scm}
v_i \leftarrow f_i(\pa_i,u_i) \in \mathcal{F}
\end{align} assign values (denoted by lowercase letters) to the respective endogenous/system variables $V_i\in\mathbf{V}$ based on the values of their parents $\Pa_i\subseteq \mathbf{V}\setminus V_i$ and the values of their respective exogenous/noise/nature variables $\mathbf{U}_i\subseteq \mathbf{U}$, and $P(\mathbf{U})$ denotes the probability function defined over $\mathbf{U}$. An intervention $\doop(\mathbf{W}), \mathbf{W} {\subset} \mathbf{V}$ on an SCM $\mathcal{M}$ occurs when (multiple) structural equations are being replaced through new non-parametric functions thus effectively creating an alternate SCM. Interventions are referred to as \emph{imperfect} if the parental relation is kept intact, as \emph{perfect} if not, and even \emph{atomic} when additionally the intervened values are being kept constant. It is important to realize that interventions are of fundamentally \emph{local} nature, and the structural equations (variables and their causes) dictate this locality. This further suggests that mechanisms remain invariant to changes in other mechanisms. An important consequence of said autonomic principles is the \emph{truncated factorization}
\begin{align} \label{eq:truncatedfactorization}
p^{\mathcal{M}}(\mathbf{v}) = \prod\nolimits_{V_i\notin \mathbf{W}} p(v_i\mid \pa_i)
\end{align}
derived by \cite{pearl2009causality}, which suggests that an intervention $\doop(\mathbf{W})$ introduces an independence of a set of intervened nodes $\mathbf{W}$ to its causal parents. For completion we mention more interesting properties of any SCM, they induce a causal graph $G$ as directed acyclic graph (DAG), they induce an observational/associational distribution denoted $p^{\mathcal{M}}$, and they can generate infinitely many interventional and counterfactual distributions using the $\doop$-operator which ``overwrites" structural equations. Note that, opposed to the Markovian SCM discussed in for instance \citep{peters2017elements}, the definition of $\mathcal{M}$ is semi-Markovian thus allowing for shared $U$ between the different $V_i$. Such a $U$ is also called \emph{hidden confounder} since it is a common cause of at least two $V_i,V_j (i\neq j)$. Opposite to that, a ``common" confounder would be a common cause from within $\mathbf{V}$. The SCM's applicability to machine learning has been shown in marketing \citep{hair2021data}, healthcare \citep{bica2020time} and education \citep{hoiles2016bounded}. As suggested by the Causal Hierarchy Theorem (CHT) \citep{bareinboim20201on}, the properties of an SCM form the Pearl Causal Hierarchy (PCH) consisting of different levels of distributions being $\mathcal{L}_1$ \emph{associational}, $\mathcal{L}_2$ \emph{interventional} and $\mathcal{L}_3$ \emph{counterfactual}. The PCH suggests that causal quantities ($\mathcal{L}_i,i\in\{2,3\}$) are in fact richer in information than statistical quantities ($\mathcal{L}_1$), and the necessity of causal information (e.g.\ structural knowledge) for inference based on lower rungs e.g.\ $\mathcal{L}_1 \not\rightarrow \mathcal{L}_2$. Finally, to query for samples of a given SCM, the structural equations are being simulated sequentially following the underlying causal structure starting from independent, exogenous variables $U_i$ and then moving along the causal hierarchy of endogenous variables $\mathbf{V}$.

{\bf Sum-Product Networks.} We follow suit with existing literature and the recent strides on tractable causal inference---mainly revolving around sum-product networks (SPN) as introduced by \cite{poon2011sum}. SPNs generalized the notion of network polynomials based on indicator variables $\lambda_{X=x}(\mathbf{x})\in[0,1]$ for (finite-state) RVs $\mathbf{X}$ from \citep{darwiche2003differential}, sum-product networks (SPN) represent a special type of probabilistic model that allows for a variety of exact and efficient inference routines. SPNs are considered as DAG consisting of product, sum and leaf (or distribution) nodes whose structure and parameterization can be efficiently learned from data to allow for efficient modelling of joint probability distributions $p(\mathbf{X})$. Formally a SPN $\mathcal{S} = (G, \mathbf{w})$ consists of non-negative parameters $\mathbf{w}$ and a DAG $G=(V,E)$ with indicator variable $\pmb{\lambda}$ leaf nodes and exclusively internal sum and product nodes given by,
\begin{align}\label{eq:spn}
    \mathsf{S}(\pmb{\lambda}) &= \sum_{\mathsf{C}\in\ch(\mathsf{S})} \mathbf{w}_{\mathsf{S},\mathsf{C}} \mathsf{C}(\pmb{\lambda}) \quad \mathsf{P}(\pmb{\lambda}) = \prod_{\mathsf{C}\in\ch(\mathsf{S})} \mathsf{C}(\pmb{\lambda}),
\end{align} where the SPN output $\mathcal{S}$ is computed at the root node ($\mathcal{S}(\pmb{\lambda})=\mathcal{S}(\mathbf{x})$) and the probability density for $\mathbf{x}$ is $p(\mathbf{x})=\frac{\mathcal{S}(\mathbf{x})}{\sum_{\mathbf{x}^{\prime}\in\mathcal{X}} \mathcal{S}(\mathbf{x}^{\prime})}$.
They are members of the family of probabilistic circuits \citep{van2019tractable}. A special class, to be precise, that satisfies properties known as completeness and decomposability. Let $\mathsf{N}$ denote a node in SPN $\mathcal{S}$, then
\begin{align}\label{eq:scope}
    \scope{\mathsf{N}} = \begin{cases}
                        \{ X\} &\text{if $\mathsf{N}$ is IV ($\lambda_{X=x}$)}\\
                        \bigcup_{\mathsf{C}\in\ch(\mathsf{N})} \scope{\mathsf{C}} &\text{else}
                        \end{cases}
\end{align} is called the scope of $\mathsf{N}$ and
\begin{align}\label{eq:spn-props}
    &\forall \mathsf{S}\in\mathcal{S}: (\forall \mathsf{C}_1,\mathsf{C}_2 \in \ch(\mathsf{S}): \scope{\mathsf{C}_1} = \scope{\mathsf{C}_2}) \\
    &\forall \mathsf{P}\in\mathcal{S}: (\forall \mathsf{C}_1,\mathsf{C}_2 \in \ch(\mathsf{S}): \mathsf{C}_1 \neq \mathsf{C}_2 \implies \scope{\mathsf{C}_1}\cap\scope{\mathsf{C}_2}=\emptyset)
\end{align} are the completeness and decomposability properties respectively. Since their introduction, SPNs have been heavily studied such as by \citep{trapp2019bayesian} that present a way to learn SPNs in a Bayesian realm whereas \citep{kalra2018online} learn SPNs in an online setting. Several different types of SPNs have also been studied such as Random SPN \citep{peharz2020random}, Credal SPNs \citep{levray2019learning} and Sum-Product-Quotient Networks \citep{sharir2018sum}) to name a few. More recently, on the intersection of machine learning and causality, \citet{zevcevic2021interventional} proposed an extension to the conditional SPN (CSPN, \citet{shao2019conditional}) capable of adhering to interventional queries. Formally, an iSPN is being defined as 
\begin{align}\label{eq:ispn}
    \mathcal{I}=(g^{\prime}_{\theta^{\prime}}:G\mapsto \Psi, \mathcal{S}^{\prime}_{\psi^{\prime}}:\mathbf{V} \mapsto [0,1])
\end{align} being a special case to the CSPN-formulation with alternate $g^{\prime}_{\theta^{\prime}}, \mathcal{S}^{\prime}_{\psi^{\prime}}$. That is, consider the general formulation of a CSPN $\mathcal{C}{=}(g_{\theta},\mathcal{S}_{\psi})$ modelling a conditional distribution $p^{\mathcal{C}}(\mathbf{Y}{\mid}\mathbf{X})$ with feed-forward neural network $g_{\theta}{:} \mathbf{X}{\mapsto} \Psi$ and SPN $\mathcal{S}_{\psi}{:} \mathbf{Y}{\mapsto} [0,1]$. By realizing that an intervention $do(\mathbf{x})$ comes with the mutilation of the causal graph $G{=}(V,E)$ such that new graph is $G^{\prime}=(V,\{(i,j):(i,j)\in E \land i\not\in\Pa_i\}$, the iSPN is able to formulate an intervention for SPN natural to the occurrence of interventions in structural causal model. The gate model $g$ orchestrates the $do$-queries such that the density estimator (SPN) can easily switch between different interventional distributions. An alternate approach to causality but also through the lens of tractability was recently considered by \citep{darwiche2021causal}.

These computational models (SPN) oppose the classical notion of semantic models (e.g.\ BNs, see Fig.\ref{fig:SPN-CI}), they trade off interpretability with efficiency. That is, an SPN might be difficult to decipher, similar to other neural methods like the multi-layer perceptron (MLP), but offer even \emph{linear} time inference---while the BN (like the Pearlian SCM) directly reasons about qualitative relationships, similar to a finger-pointing child, but at \emph{exponential} cost.

\section{Inference in Non-Causal (or Correlation-Based) Models} \label{sec:three}
\begin{figure}[t]
\centering
\includegraphics[width=.75\textwidth]{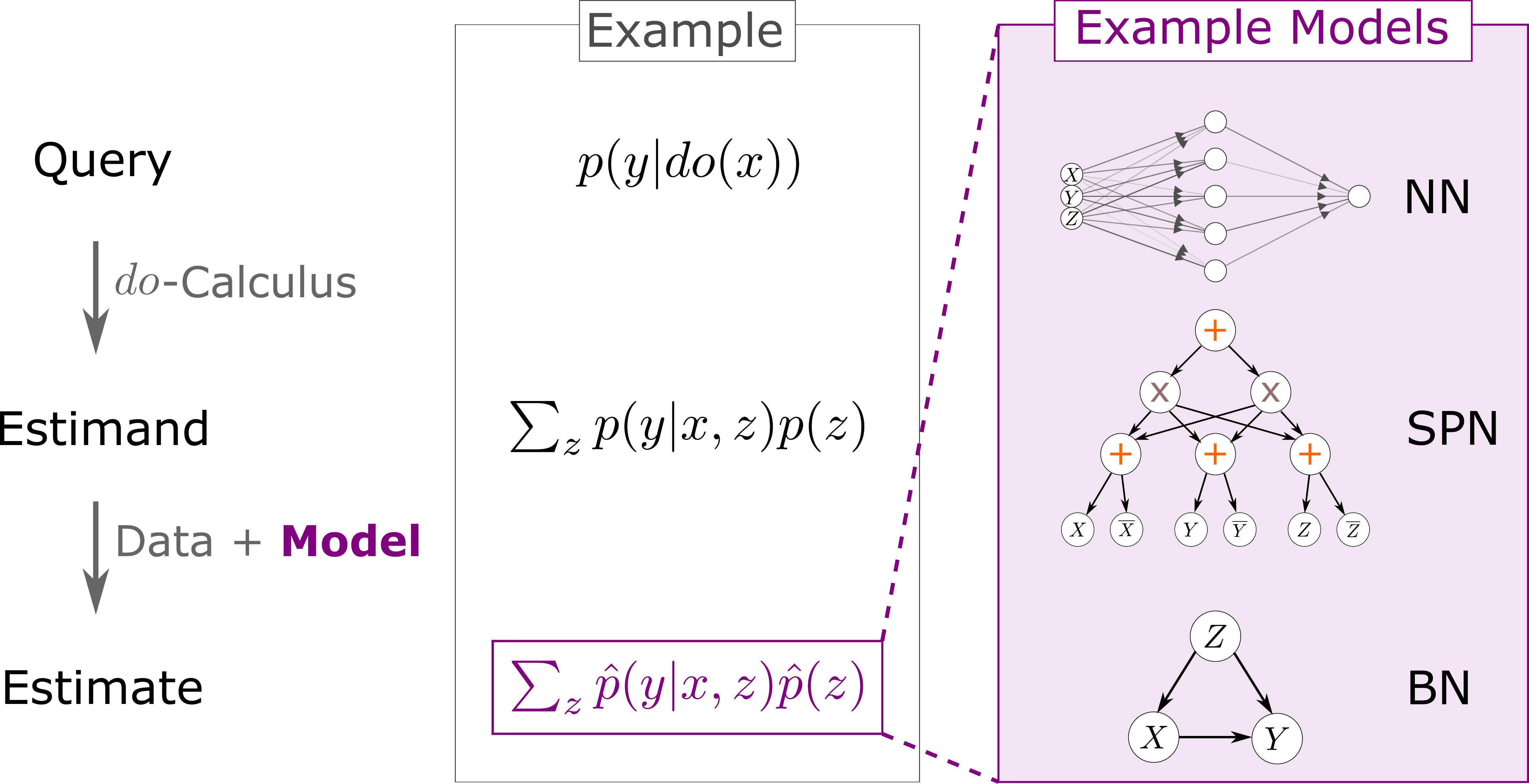}
\caption{\textbf{``Naïve" Causal Inference Schematic.} For any causal \emph{query} we could use the $\doop$-calculus \citep{pearl2009causality} to identify a statistical \emph{estimand} (grey), for which there exists a model-based \emph{estimate} (purple). Various model choices are available e.g.\ a NN, an SPN or a BN. (Best viewed in color.)}
\label{fig:SPN-CI}
\end{figure}

To expand further on the boundaries of the integration between causality and machine learning, we first perform an inspection on how causal inference can occur with correlation-based models. Fig.\ref{fig:SPN-CI} schematizes the basic, ``naïve" approach to classical causal inference that we investigate in this section. One takes the $\doop$-calculus to perform the actual ``causal'' inference, and then takes the available observational data and a model of choice (e.g.\ NN/MLP, SPN, BN) to acquire the actual estimate of the query of interest. More specifically, we will focus on SPN from the previous section, since they come with guarantees regarding probabilistic reasoning (opposed to e.g.\ MLPs) and guarantees regarding their inference tractability (opposed to e.g.\ BNs). This investigation is important since assuming the wrong causal structure or ignoring it altogether could be fatal w.r.t.\ any form of generalization out of data support as suggested in \citep{peters2017elements}. Central to said (assumed) causality is the concept of intervention. Although being a wrong statement as suggested by results on identifiability, the famous motto of Peter Holland and Don Rubin \emph{``No causation without manipulation"} \citep{holland1986statistics} phrases interventions as the core concept in causality. In agreement with this view that distributional changes present in the data due to experimental circumstances need be accounted for, we focus our analysis on queries $Q=p(\mathbf{y}{\mid}\doop(\mathbf{x}))$ with $(\mathbf{x},\mathbf{y})\in\text{Val}(\mathbf{X})\times\text{Val}(\mathbf{Y}), \mathbf{X},\mathbf{Y}\subset \mathbf{V}$ respectively. $Q$ lies on the second (interventional) level $\mathcal{L}_2$ of the PCH \citep{pearl2018book,bareinboim20201on}.

We first define the concept of a statistical estimand ($\mathcal{L}_1$) for SPN as the application of the rules of probability theory (and Bayes Theorem) to the induced joint distribution.
\begin{defin}\label{def:spn-est}
An SPN-estimand is any aggregation in terms of sums or products of conditionals, $p(\mathbf{x}{\mid}\mathbf{y})$, and marginals, $p(\mathbf{x})$, where $p(\mathbf{v})\in\mathcal{L}_1$ is the joint distribution of SPN $\mathcal{S}$ and $\mathbf{X},\mathbf{Y}\subset \mathbf{V}$ respectively.
\end{defin}
Note that for a general SPN-estimand to actually be estimable from data, full probabilistic support ($\forall \mathbf{x}: p(\mathbf{x})>0$) needs to be assumed since otherwise the estimate might be undefined. Our first insight guarantees us that we can do causal inference with SPN-estimands as depicted in Fig.\ref{fig:SPN-CI}.
\begin{prop}\label{prop:spn-est}
Let $Q\in\mathcal{L}_2$ be an identifiable query (that is, $Q$ can be purely written in $\mathcal{L}_1$ terms). There exists an SPN-estimand that answers $Q$. \hfill $\blacksquare$
\end{prop}
All the proofs not provided in the main text are found in the appendix following the references. Since SPN will act as our estimation model, it turns out that any interventional query derived from a Markovian SCM can be modelled in terms of statistical terms represented by the SCM. Due to hidden confounding, this guarantee does not hold in semi-Markovian models. Prop.\ref{prop:spn-est} ultimately suggests that inter-layer inference from $\mathcal{L}_1$ to $\mathcal{L}_2$ remains intact when choosing SPN as means of parameterization. A simple but important realization thereof is that the $\doop$-calculus \citep{pearl2009causality} can be used as the identification tool for SPN-based causal inference. While unsurprising from a causal viewpoint, from the perspective of tractable models research the result in Prop.\ref{prop:spn-est} provides a new incentive for research on the integration of both fields. A possible explanation for this surprising observation is the skeptical view by \citet{papantonis2020interventions}. They considered the usage of the SPN-BN compilation method from \citep{zhao2015relationship} for causal inference within SPN that failed due to the resulting BN being a bipartite graph in which the variables of interest were not connected (connectivity being crucial to non-trivial causal inference). Before investigating the this tractability of causal inference issue, let's define formally what we mean by \emph{tractable} inference.
\begin{defin}\label{def:tractability}
Let $R$ denote the variable in which the model's runtime scales (e.g.\ the number of edges in the DAG for an SPN, the number of variables in the DAG for a BN). A scaling of $\mathcal{O}(\text{poly}(R))$ of polynomial time is called tractable. 
\end{defin}
Note that $\emph{\text{poly}}$ includes high-degree polynomials (e.g.\ $x^{3004}$) and that for SPN we usually have $\emph{\text{poly}}:=x$, that is, linear time complexity. It is also important to note that $R$ is different for different models, but interestingly, the number of edges for SPNs does not ``explode" exponentially---so indeed, SPNs are far more efficient computation-wise even in practice. To reap initial rewards, we now prove that causal inference with SPN is in fact tractable.
\begin{corollary}\label{cor:tci-spn}
Let $Q\in\mathcal{L}_2$ be an identifiable query, $|Q|$ be its number of aggregating terms in $\mathcal{L}_1$ and $R$ be the number of edges in the DAG of SPN $\mathcal{S}$. If $|Q|<R$, then $Q$ is tractable.\hfill $\blacksquare$
\end{corollary}
Opposed to (Causal) BN where inference is generally intractable (\#P complexity), Cor.\ref{cor:tci-spn} suggests that any estimand can be computed efficiently using SPN even if the estimand identifies an interventional quantity, thereby transferring tractability of inference also to causal inference.

\section{Inference in Partially Causal Models}
An important restriction of SPN-based causal inference is that the joint distribution $p(\mathbf{v})$ of SPN $\mathcal{S}$ optimizes \emph{all} possibly derivable distributions, thereby diminishing single distribution \emph{expressivity}. That is, how ``easily" and how precisely $\mathcal{S}$ can in fact approximate our underlying distribution with $p(\mathbf{v})$. Returning to causal inference, we observe that any causal inference will hold but actual estimation from data will suffer in quality as a consequence thereof. In addition, violations of the positive support assumption might render some practical inference \emph{undefined}. Therefore, in the following we extend our analysis to \emph{partially causal models}, models that actually extend the SPN model class itself to cope with causality. More specifically, we consider interventional SPN (iSPN) firstly introduced by \citep{zevcevic2021interventional}. Our first observation is that the iSPN allows for a compressed model description over the SCM, while trading in expressivity since the iSPN has no means of computing the highest level of the PCH $\mathcal{L}_3$ being counterfactuals. The iSPN (Eq.\ref{eq:ispn}) is more ``powerful" than the SPN by construction, we state formally.
\begin{prop}\label{prop:ispntrans}
Let $E$ denote an SPN-estimand (Def.\ref{def:spn-est}). There exists a graph $G$ for which the SPN-estimand of iSPN evaluated at $G$ is $E$.
\end{prop}
\begin{proof}
There exist always an SCM $\mathcal{M}$ with induced graph $G$ such that the observational distribution of $\mathcal{M}$ and SPN-estimand $E$ correspond accordingly. Since iSPN extend regular SPN via arbitrary causal graphs, simply select $G$ as the graph of choice.
\end{proof}
Prop.\ref{prop:ispntrans} further suggests that iSPN are also joint density estimators, although being defined as a special case of conditional estimators (CSPN), and that any SPN will be covered by the observational distribution ($\mathcal{L}_1$) of a corresponding iSPN. In the following, assuming corresponding data $\mathbf{D}_i\sim p_i{\in}\mathcal{L}_2$, we prove that iSPN allow for direct causal estimation of the interventional query ($\mathcal{L}_2$). This sets iSPN apart from the setting in the previous section with regular SPN that were dependent on a ``causal inference enginge" such as Pearl's $\doop$-calculus. To illustrate the difference between the approaches, consider the following example,
\begin{align*}
    &\mathcal{M}:=(\{f_X(Z,U_X), f_Y(X,Z,U_Y), f_Z(U_Z)\},p(\mathbf{U})),
\end{align*} 
and we try inferring the query $Q$ defined as
\begin{align*}
    p(y{\mid}\doop(x))=\textstyle\sum_z p(y{\mid}x,z)p(z),
\end{align*}
where the identification equality is given by the backdoor-adjustment formula on $\mathcal{M}$ \citep{pearl2009causality,peters2017elements}. The l.h.s.\ will be modelled by an iSPN, while the r.h.s.\ consisting of multiple terms will be modelled by the SPN and \emph{required} the backdoor-adjusment formula.

An important consequence of modelling the l.h.s.\ is that the shortcomings of single distribution expressivity and positive support are being resolved. What might come as a surprise is that, although we overcome previous shortcomings and dependencies, we do not loose tractability. We observe:
\begin{prop}\label{prop:tci-ispn}\textbf{(TCI with iSPN.)}
Let $\{Q_i\}\in\mathcal{L}_2$ be a set of queries with $i\in I \subset\mathbb{N}$ and $R$ (as in Cor.\ref{cor:tci-spn}) for iSPN $\mathcal{I}$. Any $\{Q_k\}_k$ with $k\in K\subseteq I$ is tractable.
\end{prop}
\begin{proof}
There is two cases to consider (1) any fixed $Q_i$ and (2) when we switch between different $\{Q_i\}_i$ for $i\in\mathbb{N}$. For (1), since any iSPN reduces to an SPN upon parameter-evaluation, we can apply Cor.\ref{cor:tci-spn} and thereby have that $\mathcal{O}(R)$. For (2), we know the iSPN \citep{zevcevic2021interventional} uses a neural net, therefore we have $\mathcal{O}(\emph{\text{poly}}(R))$.
\end{proof}
Prop.\ref{prop:tci-ispn} seems to suggest that iSPN should always be the first choice as they don't seem to compromise while being superior to the the non-causal models that rely on causal inference engines, however, the iSPN from \citep{zevcevic2021interventional} comes with strict assumptions. Specifically, we don't assume a causal inference engine but we \emph{require interventional data}---which in a lot of practical settings is too restrictive. Also, in the case of semi-Markovian models, the iSPN falls short. Further, we observe a ``switching" restriction, which the non-causal models did not have. That is, when we have to consider multiple interventional distributions the cost will scale w.r.t.\ to the size of the gate model (either quadratically or cubically for standard feed-forward neural networks). 

\section{Inference in Structural Causal Models}\label{sec:five}
In the previous sections we discussed non-causal and partially causal models (based on SPN), showing that they are tractable---mostly trading on certain aspects such as assumptions on the data and how many queries we ask for. Although they have tractable causal inference, these methods actually lack in terms of \emph{causal expressivity}. All our previous observations were restricted to queries from $\mathcal{L}_2$, that is, interventional distributions. Why not counterfactuals? Since these models are not \emph{Structural Causal Models}. The Pearlian SCM extended historically the notion of Causal BNs (CBN) by providing both modelling capabilities for \emph{counterfactuals} but also \emph{hidden confounders}.

Now, in the following, we will move onto this more general class of models that is fully expressive in terms of the PCH. For this, consider a recent stride in neural-causal based methods ignited by the theoretical findings in \citep{xia2021causal}, where the authors introduced a \emph{parameterized} SCM. Since this term was never coined in the general sense, we provide it here for convenience.
\begin{defin}\label{def:parscm}
Let $\mathcal{M}{=}\inner{\mathbf{U},\mathbf{V},\mathcal{F},P(\mathbf{U})}$ be an SCM. We call $\mathcal{M}$ parameterized if for any $f\in\mathcal{F}$ we have that $f(\cdot,\theta)$ where $\theta$ are model parameters.
\end{defin}
In the case of \citep{xia2021causal}, the $f$ were chosen to be neural nets. Note that Def.\ref{def:parscm} actually allows for different function approximators (e.g.\ a combination of neural nets and other models), however, so far in the literature we usually have $\mathcal{F}$ be only of one such model choice. It is further important to note that any \emph{parameterized} SCM is in fact an SCM---so, an NCM is a valid SCM, furthermore, it implies the complete PCH. 

Since SCMs extended CBNs, and since CBNs are not computation graphs (like an SPN is) but rather a semantic graphs, we might conclude to inherit CBN properties when it comes to inference. Unfortunately, it turns out, this heritage of a parameterized SCM is valid and leads to their intractability for causal (marginal) inference.
\begin{theorem}\label{thm:tci-scm}
Causal (marginal) inference in parameterized SCM is NP-hard.
\end{theorem}
\begin{proof}
The simple proof might only involve reasoning that SCM extend BN such that any inferred solution for any given causal query holds in a corresponding BN if and only if it holds in the SCM because then we simply apply \citep{cooper1990computational} which showed that BN inference is NP-hard, thus also SCM inference. To write down the full argument, we apply the same technique of 3-SAT reduction. In the first step, we require a mapping between clauses from 3-SAT and SCM. As a reminder, in 3-SAT we talk of literals $Q_1,...,Q_n$ for $n\in\mathbb{N}$ with $Q_i\in\{0,1\}$ and clauses of 3 literals each $C_1,...,C_m$ for $m\in\mathbb{N}$ with $C_i(Q_k,Q_j,Q_l)$ for $k,j,l\in\{1,...,n\}$. The goal is then to find a configuration of literals $(q_1,...,q_n)$ such that each clause in the set of all clauses $C = \{C_i\}_{i=1}^m$ evaluates to 1 (read, ``true"). We know since \citep{cook1971complexity} that 3-SAT is NP-hard. So it suffices to show a reduction from 3-SAT to SCM inference i.e., that SCM inference is ``at least as hard" as 3-SAT or that an oracle of the latter would subsequently solve the former. Let $\mathcal{M}{=}\inner{\mathbf{U},\mathbf{V},\mathcal{F},P(\mathbf{U})}$ be our SCM. We do the following mapping: (1) each literal $Q_i$ will be in $\mathbf{V}$ and only depend on its ``nature" term in $\mathbf{U}$, so for each $Q_i \leftarrow f_{Q_i}(U_{Q_i}) = U_i$ where the $U_i=\mathcal{B}(\frac{1}{2})$ are random coin flips (2) each clause $C_i$ will be an effect of its causes $Q_i$, so for each $C_i \leftarrow f_{C_i}(Q_k, Q_j, Q_l, U_{C_i})$ such that $f_{C_i}$ is an indicator function of the clause. Since all clauses in $C$ need be satisfied, we create a reversed, binary tree denoted by $A_i \leftarrow f_{A_i}(\Pa_{A_i}, U_{A_i})$ (where at the leaves we have $\Pa_{A_i}=\{C_a,C_b\}$ for arbitrary two clauses and for internal nodes $\Pa_{A_i}=\{A_{i-1},C_c\}$ for some arbitrary clause $C_c$). Finally, we have $X \leftarrow f_X(A_{m-2},U_X)$ (note $m-2$ since we had $m$ clauses). This completes the mapping, the second step is to show equivalence of $p(X=1)$ to the satisfaction of $C$. Our construction implies $p(X=1)\geq p(X=1|C_s)p(C_s|U_s)p(U_s)$, where $U_s$ denotes ``true" for variables in $U$ satisfying every clause in $C$ and $C_s$ correspondingly, and we have $p(X=1|C_s)=1$, $p(C_s|U_s)$, $p(U_s)=(\frac{1}{2})^n$ so $p(X=1)>0$ when $C$ is satisfiable. If $C$ is not satisfiable, then there must be a term $p(X=1|C_q)$ for some $q\in\{1,...,m\}$ such that $p(X=1|C_q)=0$. We have $p(X=1)$ iff.\ $C$ is satisfiable. Since 3-SAT is NP-hard, we have that causal, marginal inference in parameterized SCM is NP-hard as well.
\end{proof}
From a computational perspective, the result in Thm.\ref{thm:tci-scm} is a protest against the original formulation of the SCM in terms of long-term suitability for next generation learning systems. Although being an arguably simple consequence of the BN-heritage of the SCM, still, Thm.\ref{thm:tci-scm} strongly advises against any efforts of using parameterized SCM for real-world impact. Even if the parameterization comes from powerful approximators like neural nets---causal inference remains notoriously intractable. However, for both the sake of completion and the interest of establishing the theoretical connection in the scope of this systematic investigation, we present for the first time a new parameterization of the NCM using SPN. In spite of Thm.\ref{thm:tci-scm}, this idea is indeed sensible since any \emph{partial} inference within the parameterized SCM might still be efficient. Effectively, the SPN could thereby still offer a more pragmatic alternative to e.g.\ a neural net since it would not necessarily compromise in terms of predictive performance (model capacity) since the functions of the structural equations often times describe reasonably \emph{simple} mechanisms, due to their local and restricted nature. That is, an SCM as a collective might be complex, but not its single components---similar to how a neuron in mammalian cortex shows simple activity, but human cognition in total is capable of highly complex decisions. Therefore, we now present the \emph{Tractable Neural Causal Model} (TNCM) formally.
\begin{defin}\label{def:tncm}
Let $\mathcal{M}{=}\inner{\mathbf{U},\mathbf{V},\mathcal{F},P(\mathbf{U})}$ be a parameterized SCM. If $\mathcal{F}$ exclusively defines SPN, then $\mathcal{M}$ is a Tractable Neural Causal Model.
\end{defin}
We favor the name \emph{neural} over \emph{structural} since SPN can (a) be viewed as a special type of neural/deep model (see \citet{vergari2019visualizing}), and (b) the term ``structural" so far seems exclusive to the general formalism of SCM and not to specific (ML) estimators.

\begin{figure}[t]
\centering
\includegraphics[width=.5\textwidth]{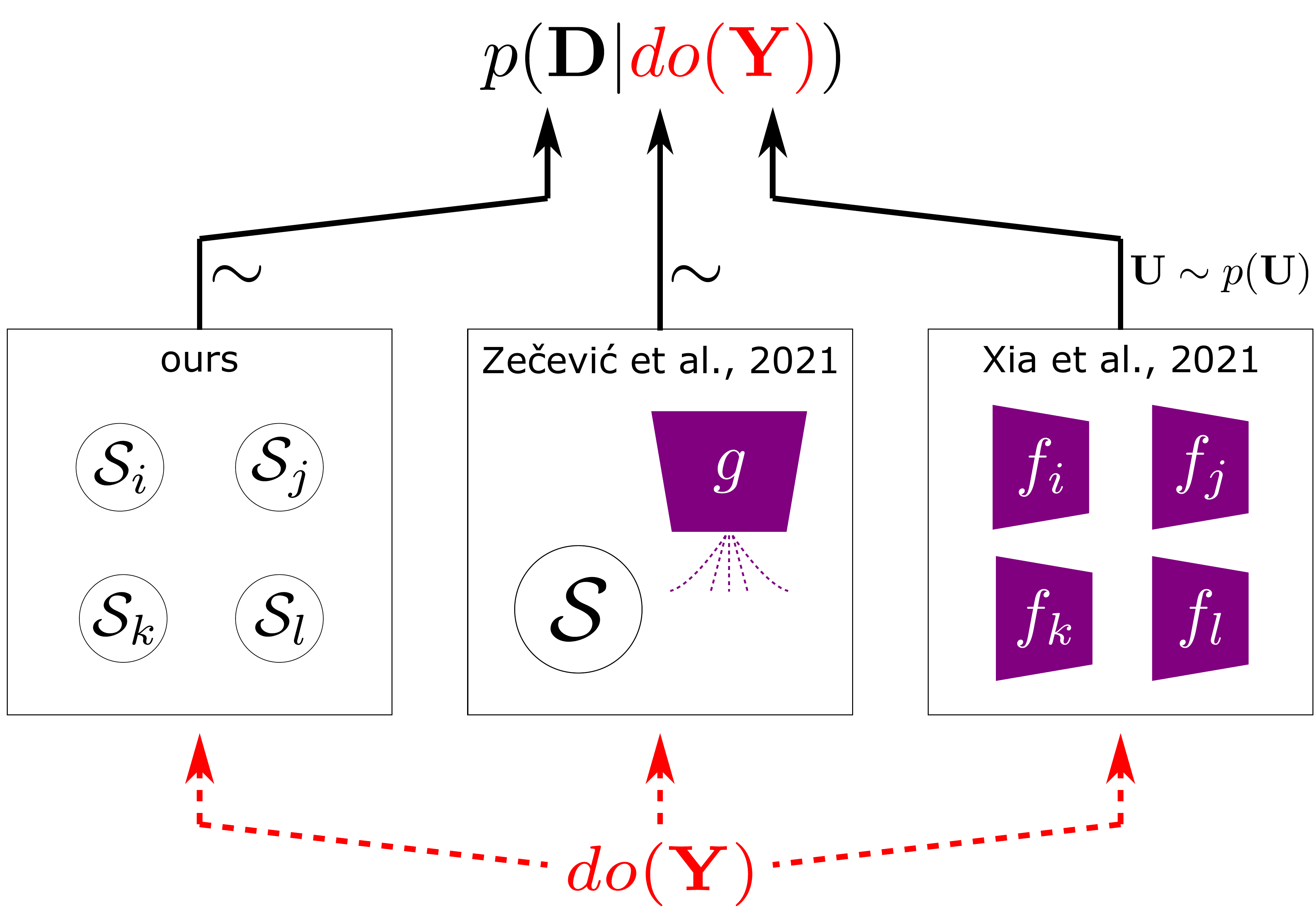}
\caption{\textbf{Schematic Overview of the Inference Processes for Selected Causal Models.} From left to right: TNCM (Def.\ref{def:tncm}), iSPN \citep{zevcevic2021interventional}, and NCM \citep{xia2021causal}. White circles denote SPN, purple quadrangles are NN. (Best viewed in color.)}
\label{fig:iSPNvsNCM}
\end{figure}

In the appendix we provide a schematic comparison of the two causal models based on SPN units i.e., the iSPN \citep{zevcevic2021interventional} from the section on partially causal models and the TNCM (Def.\ref{def:tncm}). See also Fig.\ref{fig:iSPN} in the appendix for an illustration. Evidently, the TNCM is concerned with a more complex model description (put blantly, it requires more models), yet because of that it becomes a causal model fully expressive in terms of the PCH as it poses a subset of the set of all SCMs. On a different note, in Fig.\ref{fig:iSPNvsNCM} we show a visual schematic on the different inference processes that additionally features NCM \citep{xia2021causal}. We now state the simple consequence of defining an SCM with SPN units instead of neural nets, which will further reveal one more advantage for preferring SPN over neural nets for parameterized SCM.
\begin{corollary}[NCM versus TNCM, informal]\label{cor:ncm}
Evaluating any structural equation for some SCM $\mathcal{M}$ is non-linear in NCM \citep{xia2021causal} and linear in TNCM.\hfill $\blacksquare$
\end{corollary}
Cor.\ref{cor:ncm} suggests that restricted causal inference (e.g.\ not general marginal inference) even with NCM is tractable but inefficient when compared to TNCM since the former has at least quadratic runtime complexity opposed to linear for the latter. Said comparison behaves the same for the iSPN, since Prop.\ref{prop:tci-ispn} suggests that (for a fixed iSPN state) any inference will also be linear. Further extending the comparison to other neural-causal models as suggested by \citep{zevcevic2021relating}, namely NCM-Type 2 and iVGAE. We find that for the NCM-Type 2 worse, cubic runtime complexity since modelling occurs on edge- opposed to structural equation level. For the iVGAE (which is a partially causal model), which is comparable to the iSPN in terms of model description, the time complexity is as bad as for the NCM. Therefore, iSPN (Eq.\ref{eq:ispn}) offer a clear advantage over other neural-causal models in terms of inference efficiency since any causal query will be answered in linear time, whereas NCM-variants and CBNs have worse time complexities. However, it is important to note that NCM-variants might offer for more expressivity in terms of the PCH. 

\textbf{Taxonomy.} Conclusively, a researcher might choose one model over the other based on the specific application of interest (e.g.\ efficacy versus expressivity). Upon investigating these various scenarios for tractable causal inference, we offer a conclusive overview of our tabular taxonomy for inferences in different model families in Tab.\ref{tab:over} including neural-causal inferences. Legend: OLS = Ordinary Least Squares, CNN = Convolutional Neural Networks, GAN = Generative Adversarial Networks, FBN = Functional Bayesian Network, iVGAE = interventional Variational Graph Autoencoder \citep{zevcevic2021relating}, ``Causal Circuits'' \citep{darwiche2021causal}, CausalGAN \citep{kocaoglu2017causalgan}, NCM \citep{xia2021causal}, Deep SCM \citep{pawlowski2020deep}.

\begin{table*}[t!]
\centering
\begin{adjustbox}{max width=\textwidth}
  \begin{tabularx}{\textwidth}{
  p{\dimexpr.2\linewidth-2\tabcolsep-1.3333\arrayrulewidth}
  p{\dimexpr.4\linewidth-2\tabcolsep-1.3333\arrayrulewidth}
  p{\dimexpr.4\linewidth-2\tabcolsep-1.3333\arrayrulewidth}
  }
      \toprule
      Non-Causal & \textcolor{violet}{Partially Causal} & Structural Causal \\
      \midrule
      {OLS, CNN, GAN, SPN, etc.\ } & \textcolor{violet}{{CBN, CausalVAE, iSPN, iVGAE, ``Causal Circuits", CausalGAN, etc.\ }} & {SCM, FBN, NCM, TNCM, DeepSCM, etc.}
  \end{tabularx}
\end{adjustbox}
\vspace{-.5cm}
\end{table*}
\begin{table*}[t!]
\centering
\begin{adjustbox}{max width=\textwidth}
    \begin{tabular}{c c c c c}
        \hline
        Model Family & PCH & Identification & Mechanism Inference & Marginal Inference \\
        \hline\hline
        OLS, CNN, GAN & $\mathcal{L}_1$ & \redcross & - & polynomial \\
        SPN & $\mathcal{L}_1$ & \redcross  & - & {\hspace{0.62cm} linear \ (Cor.\ref{cor:tci-spn})} \\
        CausalVAE, iVGAE, CausalGAN & $\mathcal{L}_2$ & \redcross & - & polynomial \\ 
        iSPN & $\mathcal{L}_2$ & \redcross & - & {\hspace{0.75cm} linear \ (Prop.\ref{prop:tci-ispn})} \\ 
        NCM, DeepSCM & $\mathcal{L}_3$ & \greencheck & {\hspace{0.95cm} polynomial \ (Cor.\ref{cor:ncm})} & {\hspace{0.68cm} intractable \ (Thm.\ref{thm:tci-scm})} \\
        TNCM & $\mathcal{L}_3$ & \greencheck & {\hspace{0.92cm} linear \ (Cor.\ref{cor:ncm})} & {\hspace{0.68cm} intractable \ (Thm.\ref{thm:tci-scm})} \\
        \hline
    \end{tabular}
\end{adjustbox}
    \caption{\textbf{Taxonomy of Inference in Causal Model Families.} \emph{Top}: the three classes perspective of non-causal, partially causal and structural causal models with known models (non-exhaustive list). \emph{Bottom}: Summarizing tractability properties discussed throughout the paper. PCH layer $\mathcal{L}_i$ with $i$ being the upper bound on causal quantities expressible (e.g.\ $i=3$ means any causal quantity according to Pearl can be generated). Identification suggests that cross-layer inferences can be performed (e.g.\ no external identification engine like $\doop$-calculus is necessary). A dash (-) denotes that the structural equations of a corresponding SCM can not even be defined in the given model family. Marginal inference refers to whether the general computation scheme $p(x)=\sum_{\mathbf{v}\setminus x} p(x,\mathbf{v})$ is computable tractably. Mechanism inference refers to the tractability of the computation of any single sub-module (i.e., structural equation).  (Best viewed in color.)}
\label{tab:over}
\end{table*}

\section{Empirical Illustration} \label{sec:four}

\paragraph{Training and Estimation with TNCM.}
Since TNCM are a special case of SCM with SPN as parameterizing units, we can apply inference in the same way. That is, we make use of the truncated factorization formula \citep{pearl2009causality} by choosing a sample (or Monte Carlo) based approximation thereof,
\begin{equation}\label{eq:TR}
    p(\mathbf{V}=\mathbf{v}{\mid}\doop(\mathbf{X}=\mathbf{x})) \approx \frac{1}{m}\textstyle\sum_i^m \textstyle\prod_{\mathbf{V}\setminus \mathbf{X}} f(\mathbf{v}, \theta),
\end{equation} where $m$ is the number of samples for the unmodelled/noise terms $U_i$ and $f$ as in Def.\ref{def:parscm}. The intuition behind this formula is that an intervention will mutilate the original causal graph deleting dependence on $\mathbf{X}$'s parents. To perform training, one can simply resort to the maximization of the probability in terms of the negative log-likelihood to account for numerical stability, that is $\theta^* = \arg\min_{\theta\in\Theta} -\frac{1}{n} \sum_i^n \log(p(\mathbf{v}{\mid}\doop(x)))$ where $n$ is the number of data points. The consistency criterion refers to the assumption that a query like $p(y=1,x=1{\mid}\doop(x=0))$ should automatically evaluate to zero since it would be inconsistent to observe $x$ as opposing the intervention.

\begin{figure*}[t!]
\centering
\includegraphics[width=\textwidth]{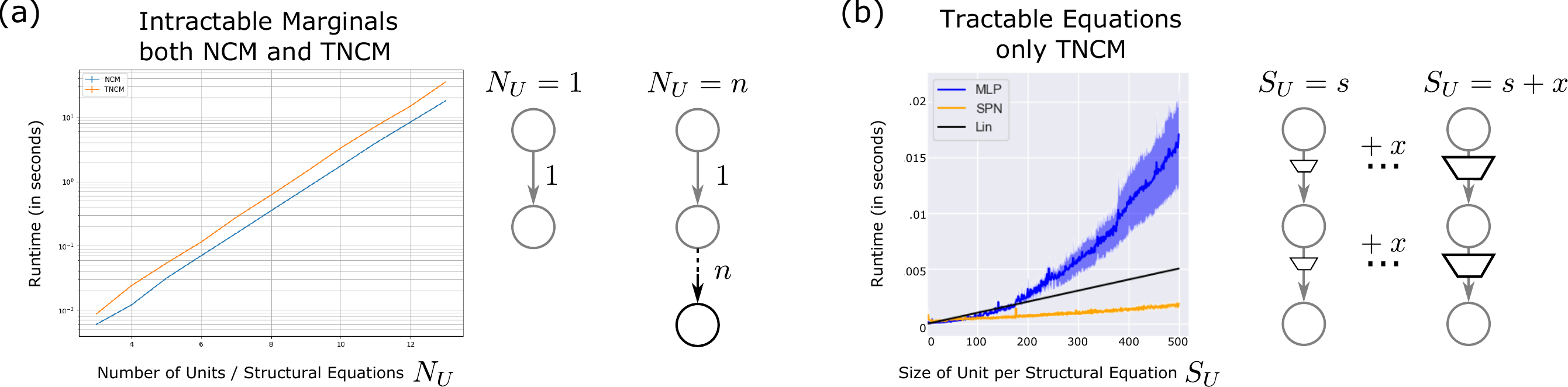}
\caption{\textbf{Experiments Showing Different ``Dimensions'' of Tractability.} (a) log-scale plot of NCM \citep{xia2021causal} versus TNCM (Def.\ref{def:tncm}) inference runtime (in seconds) when increasing the size of the SCM i.e., the number of causal variables and thereby both the number of structural equations and subsequent models to be trained. We observe intractability as predicted by Thm.\ref{thm:tci-scm}. (b) regular-scale plot of runtimes for TNCM and NCM when increasing the capacity of the models to be learned for each of the structural equation. We observe linear tractability only for TNCM. (Best viewed in color.)}
\label{fig:tract}
\end{figure*}

We investigate the newly-introduced TNCM (Def.\ref{def:tncm}) specifically. We first ``sanity check" the model by checking for causal effect and general density estimation. Then we conduct two experiments regarding tractability of causal inference. More specifically, we answer the following questions:
\begin{itemize}
    \item[\textbf{Q1.}] To which degree are causal effects being captured on qualitatively different structures?
    \item[\textbf{Q2.}] How is the estimation quality for interventional distribution modelling?
    \item[\textbf{Q3.}] How does time complexity scale when increasing the SCM's size, that is, number of modelling units ($N_U$)?
    \item[\textbf{Q4.}] How does time complexity scale when increasing the size of each unit per SCM structural equation ($S_U$)?
\end{itemize}

Due to space constraints we are restricted in presentation for the main text and choose to focus on the most difficult, important questions of \textbf{Q3} and \textbf{Q4}. For the rest, and also details regarding our synthetic data sets, the protocol, and the parameters---we refer to the appendix \ref{app:one}.

For completition's sake, we briefly mention that both \textbf{Q1} and \textbf{Q2} are answered in favor of TNCM (Fig.\ref{fig:ate} and Tab.\ref{tab:jsd}), that is, both causal effect estimation as well as general density estimation are competitive with the NCM \citep{xia2021causal}--while TNCM reaps the benefits discussed throughout Sec.\ref{sec:five}.

\textbf{Q3.} \emph{Increasing $N_U$.} Consider Fig.\ref{fig:tract}, Left. We increase the size of the system which is arguably the most common form of scaling and relevant to the development of e.g.\ complex social networks or for biomedical analysis of complex proteins. As predicted by our intractability result in Thm.\ref{thm:tci-scm} both NCM \citep{xia2021causal} and TNCM (Def.\ref{def:tncm}) scale \emph{exponentially}, since they are both parameterized SCM that do not represent computational but rather semantic relations of the variables. The offset difference stems from the specifics of our experimental setup and is negligible.

\textbf{Q4.} \emph{Increasing $S_U$.} Consider Fig.\ref{fig:tract}, Right. We increase the size of each of the system's models (each structural equation) with the reasoning that in nature it might occur for a causal relation to be notoriously complex, for instance again in the medical domain the causal mechanism that revolves around risks of smoking as long-standing example \citep{pearl2018book}--although one might argue that at a more fine-grained view of the mechanisms they might again become simple. As suggested by our simple corollary (Cor.\ref{cor:ncm}), only the TNCM (Def.\ref{def:tncm}) is linear tractable.

\section{Concluding Discussion and Future Directions}
Throughout this work we first coined the terms to define a spectrum of causal models (``non-causal", ``partially causal" and ``structural causal") to then investigate their tractability properties when performing causal inferences. We highlighted the importance of tractability for long-term development of practical, next-generation learning systems while providing a broad, comprehensive overview of the discriminative properties for the existing ``Zoo" of models (see our taxonomy in Tab.\ref{tab:over}). Establishing this overview involved, among other things, that we proved the general intractability result for parameterized SCM (see Thm.\ref{thm:tci-scm})--which we did using the classical techniques previously used for belief networks. However, we also showed ways of coping with intractability---by proposing the TNCM which we demonstrated theoretically and empirically---to pave the way for future research. 

We believe that future models will fall into the defined spectrum of causal models and with them their tractability properties. Since future models will require both number-wise more interactions and also more complex interactions, research at the integration of causality and AI/ML will inevitably encounter the tractability question. Causal inference engines like the $\doop$-calculus are a tremendously powerful tool, yet ultimately, \emph{complete automation} is what AI seems to aim for. As suggested in Tab.\ref{tab:over}, making partially causal models like iSPN \citep{zevcevic2021interventional} ``less partial" or structural causal models like NCM \citep{xia2021causal} ``more tractable" both aim at the same end result---tractable causal models. Coming from the tractability perspective, the original introduction of SPNs from ACs \citep{darwiche2003differential,poon2011sum}---that allowed for replacing semantic relations through computational ones---might provide hints for a tractable view onto the Pearlian notion of causality. Also, providing a large-scale example akin to ImageNet \citep{krizhevsky2012imagenet}, might be beneficial for future investment into tractable causal inference research.

\textbf{Final Remarks.} We hope that our impossibility result alongside the taxonomy for tractability in causal models (and the initial model, TNCM, we deduced from it) can raise awareness for this novel research direction since achieving success with causality in real world downstream tasks will not only depend on learning correct models as we also require having the practical ability to gain access in finite resources to model inferences, ideally as efficiently as possible.

\clearpage 
\section*{Acknowledgements}
The authors acknowledge the support of the German Science Foundation (DFG) project “Causality, Argumentation, and Machine Learning” (CAML2, KE 1686/3-2) of the SPP 1999 “Robust Argumentation Machines” (RATIO). This work was supported by the ICT-48 Network of AI Research Excellence Center “TAILOR” (EU Horizon 2020, GA No 952215), the Nexplore Collaboration Lab “AI in Construction” (AICO) and by the Federal Ministry of Education and Research (BMBF; project “PlexPlain”, FKZ 01IS19081). It benefited from the Hessian research priority programme LOEWE within the project WhiteBox \& the HMWK cluster project “The Third Wave of AI” (3AI).
\bibliography{tncm}


\clearpage
\appendix
\section{Paper: ``A Taxonomy for Inference in Causal Model Families''}
We make use of this appendix following the main paper to provide the experimental details, an inference algorithm, additional results regarding causal effect and density estimation, all the plots and the link to our open code base.

\subsection{Difference between iSPN and TNCM}
\begin{figure*}[h!]
    \centering
    \includegraphics[width=.95\textwidth]{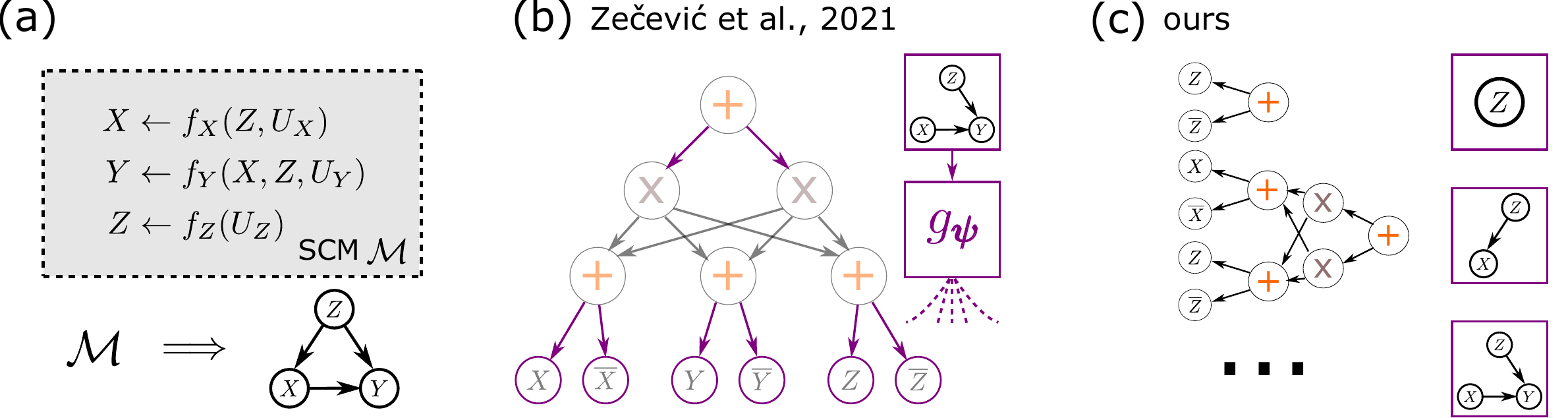}
    \caption{\textbf{Tractable Causal Inference Models.} (a) The unobserved SCM $\mathcal{M}$ implies a causal graph $G$ and generates the data to be used for estimation, (b) shows an iSPN \citep{zevcevic2021interventional} that uses a gate model to estimate causal effects, whereas (c) is a TNCM (Def.\ref{def:tncm})---a partially tractable approximation to $\mathcal{M}$. (Best viewed in color.)}
    \label{fig:iSPN}
\end{figure*}

In Fig.\ref{fig:iSPN} we provide a schematic comparison of the two causal models based on SPN units i.e., the iSPN \citep{zevcevic2021interventional} from the section on partially causal models and the TNCM (Def.\ref{def:tncm}) that we introduce in this paper as a stepping stone towards tractable equivalents of structural causal models. Both model families use SPN modules but in a completely different way. Evidently, the TNCM is concerned with a more complex model description (put blantly, it requires more models)--yet because of that--it becomes a causal model fully expressive in terms of the PCH as it poses a subset of the set of all SCMs.

\setcounter{prop}{0}
\setcounter{corollary}{0}
\subsection{Proofs for Proposition 1 and Corollaries 1, 2}
For ease of presentation, the following proofs were not shown in the main text. They are now provided here within the appendix sections.
\begin{prop}\label{prop:spn-est}
Let $Q\in\mathcal{L}_2$ be an identifiable query (that is, $Q$ can be purely written in $\mathcal{L}_1$ terms). There exists an SPN-estimand that answers $Q$.
\end{prop}
\begin{proof}
Under positive support assumption, trivially follows from SPN being universal density approximators \citep{poon2011sum} and $Q$ being identifiable.
\end{proof}
\begin{corollary}\label{cor:tci-spn}
Let $Q\in\mathcal{L}_2$ be an identifiable query, $|Q|$ be its number of aggregating terms in $\mathcal{L}_1$ and $R$ be the number of edges in the DAG of SPN $\mathcal{S}$. If $|Q|<R$, then $Q$ is tractable.
\end{corollary}
\begin{proof}
From \citep{poon2011sum} we have that $\mathcal{S}$ does a single term, bottom-up computation linearly in $R$. Through Prop.\ref{prop:spn-est} and $|Q|<R$ it follows that $\mathcal{O}(R)$.
\end{proof}
\begin{corollary}[NCM versus TNCM, informal]\label{cor:ncm}
Evaluating any structural equation for some SCM $\mathcal{M}$ is non-linear in NCM \citep{xia2021causal} and linear in TNCM.
\end{corollary}
\begin{proof}
The forward pass for the neural nets in NCM has polynomial runtime complexity and will specifically depend on the architecture, generally lower bounded by a quadratic term. For SPN, as previously established, generally linear.
\end{proof}

\subsection{Experimental Details, Inference Algorithm and Additional Results}\label{app:one}

{\bf Data Sets.} Since we are interested in qualitative behavior in light of the theoretical results established previously, we consider custom SCM simulations. For instance consider the following two models: the collider SCM given by
\begin{align*}
\mathcal{M}_1 &= 
\left\{\begin{aligned}\quad
X & \leftarrow & f_X(W, U_X) = & W\wedge U_X\\
Y & \leftarrow & f_Y(U_Y) = & U_Y\\
Z & \leftarrow & f_Z(X, Y, U_Z) = & X\vee (Y\wedge U_Z)\\
W & \leftarrow & f_W(U_W) = & U_W
     \end{aligned}\right.
\end{align*} 
and a simple chain SCM which has no confounding given by
\begin{align*}
\hspace{-.8cm}
\mathcal{M}_2 &= 
\left\{\begin{aligned}\quad
X & \leftarrow & f_X(U_X) = & U_X\\
Y & \leftarrow & f_Y(X, U_Y) = & X\wedge U_Y\\
Z & \leftarrow & f_Z(Y, U_Z) = & Y\wedge U_Z\\
W & \leftarrow & f_W(Z, U_W) = & Z\wedge U_W,\\
     \end{aligned}\right.
\end{align*} 
 and the confounded SCM is given by
\begin{align*}
\hspace{1.4cm}
\mathcal{M}_3 &= 
\left\{\begin{aligned}\quad
X & \leftarrow & f_X(Z, U_X) = & Z\vee U_X\\
Y & \leftarrow & f_Y(X,Z, U_Y) = & (X\wedge U_Y)\oplus (Z\wedge U_Y)\\
Z & \leftarrow & f_Z(U_Z) = & U_Z\\
W & \leftarrow & f_W(X, U_W) = & X\wedge U_W,\\
     \end{aligned}\right.
\end{align*}
and the backdoor SCM given by
\begin{align*}
\hspace{.4cm}
\mathcal{M}_4 &= 
\left\{\begin{aligned}\quad
X & \leftarrow & f_X(Z, U_X) = & Z \oplus U_X\\
Y & \leftarrow & f_Y(W, X, U_Y) = & X\wedge (W\wedge U_Y)\\
Z & \leftarrow & f_Z(U_Z) = & U_Z\\
W & \leftarrow & f_W(Z, U_W) = & Z\wedge U_W,
     \end{aligned}\right.
\end{align*}
where $\oplus,\vee,\wedge$ denote logical XOR, OR, and AND. Note that (for simplicity of analysis) we consider binary variables, however, (T)NCM naturally extend to the categorical and continuous variables. Note that the collider is an unconfounded structure, thereby conditioning amounts to intervening, $p(y{\mid}x)=p(y{\mid}\doop(x))$, while for the backdoor this equality does not hold - thus the causal effect from $X$ on $Y$ is confounded via the backdoor $X\leftarrow ...$ over nodes $Z,W$. We choose $U\sim \text{Unif}(a,b)$ to be uniform random variables each, and we randomize parameters $a,b$.

{\bf Protocol and Parameters.} To account for reproducibility and stability of the presented results, we used learned models for four different random seeds each parameterization of any given underlying SCM. For the NCM's neural networks, we deploy simple MLP with three hidden layers of 10 neurons each, and the input-/output-layers are $|\Pa_i|+1$ and 1 respectively. For the TNCM's SPNs, we deploy simple two-layer SPNs (following the layerwise principle introduced in \citet{peharz2020einsum}) where the first layer consists of leaf nodes, the second layer of product nodes, the third layer of sum nodes and a final product node aggregation. The number of channels is set to 30. We use ADAM \citep{kingma2014adam} optimization, and train up to three passes of 10 k data points sampled from the observational distribution of any SCM. For experiments in which the size of the SCM is being increased, we use a simple chain and extend it iteratively. For experiments in which the capacity of the mechanism (or units) of the parameterized SCM are being increased, we use a fixed chain SCM structure and scale the model capacity linearly. I.e., the MLPs increase their hidden layers neurons number while SPNs increase their layer channel. For causal effect estimation, we focus on the average treatment effect given by $ATE(T,E):=\mathbb{E}[E{\mid}\doop(T=1)]-\mathbb{E}[E{\mid}\doop(T=1)]$ that for the binary setting reduces to probabilistic difference $p(Y=1{\mid}\doop(X=1))-p(Y=1{\mid}\doop(X=0))=ATE(T,E)$. For measuring density estimation quality, we resort to the Jensen-Shannon-Divergence (JSD) with base 2 that is bounded in $[0,1]$ where 0 indicates identical probability density functions i.e., an optimal match in terms of JSD.

\subsection{Q1 ATE Estimation, see Fig.\ref{fig:ate}} We observe adequate modelling of the ATEs in both neural-causal models. The worst score on ATE for this binary setting would be 2, while the observed values are in the range $[0,0.09]$ thus significantly less. The confounded cases ($\mathcal{M}_{3/4}$) are indeed inferred correctly. TNCM with chosen hyperparameters achieves sligthly worse score than the NCM but with the tendency of reduced variance in the estimates. We argue that the observed variances stem from the choice of SCM parameterizations.

\begin{figure}[t!]
  \begin{minipage}[t!]{.45\linewidth}
    \centering
    \includegraphics[scale=.3]{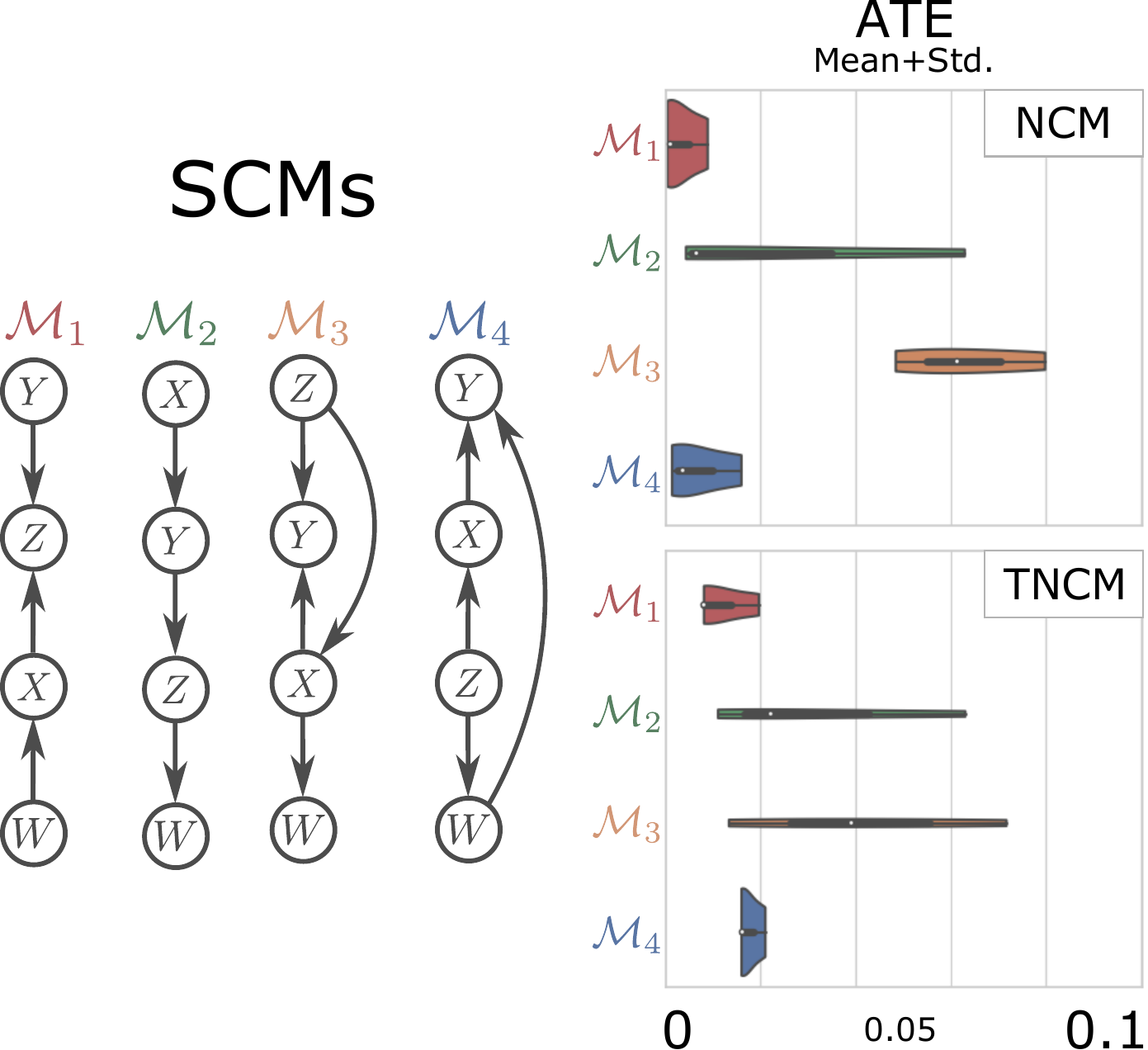}%
    \caption
      {%
        \textbf{ATE.} Averaged over multiple random seeds over multiple paramterizations of the given SCM. Both NCM and TNCM perform well in estimating causal effects. (Best viewed in color.)%
        \label{fig:ate}%
      }%
  \end{minipage}\hfill
  \begin{minipage}[t!]{.45\linewidth}
    \centering
    \vspace{-3.8cm}
    \begin{tabular}{cc|ccc}
    \hline
    & SCM & $\mathcal{L}_1$ & $\mathcal{L}^{\doop(X=0)}_2$ & $\mathcal{L}^{\doop(X=1)}_2$ \\
    \hline
    \multirow{4}{*}{\rotatebox[origin=c]{90}{NCM}} &$\mathcal{M}_1$&.011&.010&.026\\ &$\mathcal{M}_2$&.017&.011&.020\\ &$\mathcal{M}_3$&.012&.009&.030\\ &$\mathcal{M}_4$&.006&.010&.005\\
    \hline
    \multirow{4}{*}{\rotatebox[origin=c]{90}{TNCM}} &$\mathcal{M}_1$&.075&.040&.310\\ &$\mathcal{M}_2$&.012&.032&.022\\ &$\mathcal{M}_3$&.032&.024&.033\\ &$\mathcal{M}_4$&.029&.021&.011\\
    \hline
    \end{tabular}
    \captionof{table}
      {%
      \textbf{Density Estimation.} Averaged JSD values on three different distributions for each of the four SCMs for both NCM and TNCM.%
        \label{tab:jsd}%
      }
  \end{minipage}
\end{figure}

\subsection{Q2 Density Estimation, see Tab.\ref{tab:jsd}} We observe adequate modelling of the different densities (the actual plots for NCM are provided \href{https://anonymous.4open.science/r/TNCM/media/Figure-App-NCM.pdf}{[CLICK HERE, NCM Plots]} and for TNCM are provided \href{https://anonymous.4open.science/r/TNCM/media/Figure-App-TNCM.pdf}{[CLICK HERE, TNCM Plots]}) since error rates lie mostly in the low single digit domain. Most notably is the increased variance of the $\doop(X=1)$ distribution for TNCM on $\mathcal{M}_1$. Observing closely, we see that even the other distributions already show less-optimal performance. Since all experiments are conducted with the same, simple architectures, we argue that this non-optimization is explanatory.

\vspace{3cm}
\subsection{Code Repository and Technical Details}
This brief section provides anchor points for any further relevant information.
\paragraph{Code Repo.} Our code repository alongside visualizations is publically available at: \url{https://anonymous.4open.science/r/TNCM/}

\paragraph{Technical Details.} All experiments are being performed on a MacBook Pro (13-inch, 2020, Four Thunderbolt 3 ports) laptop running a 2,3 GHz Quad-Core Intel Core i7 CPU with a 16 GB 3733 MHz LPDDR4X RAM on time scales between a few second and an hour.


\end{document}